%% file: paper.tex
\documentclass{article}
\usepackage{a4wide}

\usepackage{graphicx}
\usepackage{color}
\usepackage{amsmath,amssymb}
\usepackage{float}
\usepackage{subfigure}

\newcommand{\argmax}{\mathrm{argmax}}
\newcommand{\sign}[1]{\mathrm{sign}\left(#1\right)}
\newcommand{\diag}[1]{\mathrm{diag}\left(#1\right)}
\newcommand{\GinvPhi}[1]{G_b^{-1}(\Phi_{0,\sigma^2}(#1))}
\newcommand{\PhiinvG}[1]{\Phi_{0,\sigma^2}^{-1}(G_b(#1))}

\author{\begin{tabular}{ccc}
    {\bf Fabian L. Wauthier} &~~~~~& {\bf Michael I. Jordan}\\
  Computer Science Division && Computer Science Division\\ 
  University of California, Berkeley && University of California, Berkeley\\
  {\tt flw@cs.berkeley.edu} && {\tt jordan@cs.berkeley.edu}
  \end{tabular}
}
\title{Heavy-Tailed Processes for Selective Shrinkage}
\date{}

\begin{document}

\maketitle

\begin{abstract}
Heavy-tailed distributions are frequently used to enhance the robustness of
regression and classification methods to outliers in output
space. Often, however, we are confronted with ``outliers'' in
input space, which are isolated observations in sparsely populated
regions. We show that heavy-tailed stochastic processes (which we
construct from Gaussian processes via a copula), can be used to
improve robustness of regression and classification estimators to such
outliers by selectively shrinking them more strongly in sparse regions
than in dense regions. We carry out a theoretical analysis to show
that selective shrinkage occurs, provided the marginals of the
heavy-tailed process have sufficiently heavy tails. The analysis is
complemented by experiments on biological data which indicate
significant improvements of estimates in sparse regions while
producing competitive results in dense regions.
\end{abstract}
Gaussian process classifiers (GPCs)~\cite{rasmussenwilliams} provide a
Bayesian approach to nonparametric classification with the key
advantage of producing predictive class probabilities. Unfortunately,
when training data are unevenly sampled in input space, GPCs tend to
overfit in the sparsely populated regions. Our work is motivated by
an application to protein folding where this presents a major
difficulty.  In particular, while Nature provides samples of protein
configurations near the global minima of free energy functions,
protein-folding algorithms necessarily explore regions far from the
minimum.  If the estimate of free energy is poor in those
sparsely-sampled regions then the algorithm has a poor guide towards
the minimum.  More generally this problem can be viewed as one of
``covariate shift,'' where the sampling pattern differs in the
training and testing phase.

In this paper we investigate a GPC-based approach that addresses
overfitting by shrinking predictive probabilities towards conservative
values. For an unevenly sampled input space it is natural to consider
a {\em selective shrinkage} strategy: we wish to shrink probability
estimates more strongly in sparse regions than in dense regions.  To
this end several approaches could be considered. If sparse regions can
be readily identified, selective shrinkage could be induced by
tailoring the Gaussian process (GP) kernel to reflect that
information. In the absence of such knowledge, Goldberg and
Williams~\cite{Goldberg98regressionwith} showed that Gaussian process
regression (GPR) can be augmented with a GP on the log noise level.
More recent work has focused on partitioning input space into discrete
regions and defining different kernel functions on each. Treed
Gaussian process regression~\cite{Gramacy07bayesiantreed} and Treed
Gaussian process classification~\cite{broderick:2009:classification}
represent advanced variations of this theme that define a prior
distribution over partitions and their respective kernel
hyperparameters.  Another line of research which could be adapted to
this problem posits that the covariate space is a nonlinear
deformation of another space on which a Gaussian process prior
is placed~\cite{damian01deform, schmidt03deform}. Instead of
directly modifying the kernel matrix, the observed non-uniformity of
measurements is interpreted as being caused by the spatial
deformation.  A difficulty with all these approaches is that posterior
inference is based on MCMC, which can be overly slow for the
large-scale problems that we aim to address.

This paper presents an alternative approach to selective shrinkage
which replaces the Gaussian process underlying GPC with a stochastic
process that has heavy-tailed marginals (e.g., Laplace, hyperbolic
secant, or Student-\(t\)). While heavy-tailed marginals are generally
viewed as providing robustness to outliers in the \emph{output space}
(i.e., the response space), the selective shrinkage notion can be
viewed as a form of robustness to outliers in the \emph{input space}
(i.e., the covariate space).  Indeed, selective shrinkage means the
data points that are far from other data points in the input space are
regularized more strongly. We provide a theoretical analysis and
empirical results to show that inference based on stochastic
processes with heavy-tailed marginals yields precisely this kind of
shrinkage.

The paper is structured as follows:
Section~\ref{sec:gpc_and_shrinkage} provides background on GPCs.  We
present a construction of heavy-tailed stochastic processes in
Section~\ref{sec:sparse_proc_and_copula_trick} and show that inference
reduces to standard computations in a Gaussian process.  An analysis
of our approach is presented in Section~\ref{sec:selective_shrinkage}
and details on inference algorithms are presented in
Section~\ref{sec:sparse_proc_classification}.  Experiments on
biological data in Section~\ref{sec:experiments} demonstrate that
heavy-tailed process classification substantially outperforms GPC in
sparse regions while performing competitively in dense regions. The
paper concludes with an overview of related research and final remarks
in Sections~\ref{sec:related} and~\ref{sec:conclusion}.
\section{Gaussian process classification and shrinkage}\label{sec:gpc_and_shrinkage}
A Gaussian process (GP)~\cite{rasmussenwilliams} is a prior on
functions \(z:\mathcal{X}\rightarrow\mathbb{R}\) defined through a
mean function (usually identically zero) and a symmetric positive
semidefinite kernel \(k(\cdot, \cdot)\). For a finite set of
locations \(X = (x_1, \ldots, x_n)\) we write \(z(X) \sim p(z(X)) =
\mathcal{N}(0, K(X, X))\) as a random variable distributed according
to the GP with finite-dimensional kernel matrix \([K(X, X)]_{i,j} =
k(x_i, x_j)\). Let \(y\) denote a vector of binary class labels
associated with measurement locations \(X\)\footnote{To improve the
  clarity of exposition, we only deal with binary classification for
  now. A full multiclass classification model will be used for
  our experiments.}. For Gaussian process classification
(GPC)~\cite{rasmussenwilliams} the probability that a test point
\(x_*\) is labeled as class \(y_* = +1\), given training data \((X, y)\), 
is computed as
\begin{align}
  p(y_* = +1|X, y, x_*) &= \mathbb{E}_{p(z(x_*)| X, y, x_*)}\left(\frac{1}{1 + \exp\{-z(x_*)\}}\right)\label{eq:gpc_predictive_prob}\\
  p(z(x_*)|X, y, x_*) &= \int p(z(x_*)|X, z(X), x_*) p(z(X)|X, y)dz(X).\nonumber
\end{align}
The predictive distribution \(p(z(x_*)|X, y, x_*)\) represents a
regression on \(z(x_*)\) with a complicated observation model
\(y|z\). From Eq.~\eqref{eq:gpc_predictive_prob} we observe that we
could selectively shrink the prediction \(p(y_* = +1|X, y, x_*)\)
towards a conservative value \(1/2\) by selectively shrinking
\(p(z(x_*)|X, y, x_*)\) closer to a point mass at zero. Our paper
takes this intuition and shows that such selective shrinkage can be
achieved by replacing the GP underlying GPC with a stochastic process
that has sufficiently heavy tails.
\section{Heavy-tailed stochastic processes via the Gaussian copula}\label{sec:sparse_proc_and_copula_trick}
In this section we construct the heavy-tailed stochastic process by
transforming a GP. As with the GP, we will treat the new process as a
prior on functions. Suppose that \(\diag{K(X, X)} = \sigma^2 {\bf
  1}\). We define the heavy-tailed process \(f(X)\) with marginal
c.d.f.\ \(G_b\) as
\begin{align}
  z(X) &\sim \mathcal{N}(0, K(X, X))\label{eq:const0}\\
  u(X) &= \Phi_{0, \sigma^2}(z(X))\label{eq:const2}\\
  f(X) &= G_b^{-1}(u(X)) = G_b^{-1}(\Phi_{0,\sigma^2}(z(X)))\nonumber        . 
\end{align}
Here the function \(\Phi_{0, \sigma^2}(\cdot)\) is the c.d.f.\ of a
centered Gaussian with variance \(\sigma^2\). Presently, we only
consider the case when \(G_b\) is the (continuous) c.d.f.\ of a
heavy-tailed density \(g_b\) with scale parameter \(b\) that is
symmetric about the origin. Examples include the Laplace, hyperbolic
secant and Student-\(t\) distribution. We note that other authors have
considered asymmetric or even discrete
distributions~\cite{Chu04gaussianprocesses,
  pitt2006gaussiancopularegression, multivariatedisp2000song} while
Snelson et al.~\cite{snelson04warped} use arbitrary monotonic
transformations in place of \(\GinvPhi{\cdot}\). The
process \(u(X)\) has the density of a Gaussian
copula~\cite{copulabook, multivariatedisp2000song} and is critical in
transferring the correlation structure encoded by \(K(X, X)\) from
\(z(X)\) to \(f(X)\). If we define \(z(f(X)) = \PhiinvG{f(X)}\), it is
well known~\cite{jaimungal09kcp, liu2009nonparanormal,
  pitt2006gaussiancopularegression, snelson04warped,
  multivariatedisp2000song} that the density of \(f(X)\) takes the form
\begin{align}
  p(f(X)) = \frac{\prod_{i=1} g_b(f(x_i))}{|K(X, X)/\sigma^2|^{1/2}}
  \exp\left\{-\frac{1}{2}z(f(X))^\top\left[K(X, X)^{-1} - \frac{I}{\sigma^2}\right]z(f(X))\right\}.\label{eq:warped_likelihood}
\end{align}
Observe that if \(K(X, X) = \sigma^2 I\) then \(p(f(X)) = \prod_{i=1}
g_b(f(x_i))\). A prior Gaussian process with
independent components induces a Heavy-tailed process with independent
components. Also note that if \(G_b\) were chosen to be Gaussian,
we would recover the Gaussian process. The predictive distribution
\(p(f(x_*)|X, f(X), x_*)\) can be interpreted as a Heavy-tailed
process regression (HPR). It is easy to see that its computation can
be reduced to standard computations in a Gaussian model by nonlinearly
transforming observations \(f(X)\) into \(z\)-space. Specifically, the
predictive distribution in \(z\)-space satisfies
\begin{align}
  &p(z(x_*)|X,f(X),x_*) = \mathcal{N}(\mu_{*}, \Sigma_{*})\label{eq:zspace_pred_start}\\
  &\mu_{*} = K(x_*, X)K(X, X)^{-1}
  z(f(X))\\
  &\Sigma_{*} = K(x_*, x_*) - K(x_*,X)K(X, X)^{-1}K(X, x_*)\label{eq:zspace_pred_end}.
\end{align}
The corresponding distribution in \(f\)-space follows by another
change of variables. Having defined the heavy-tailed stochastic
process in general we now turn to analyze its shrinkage properties.
\section{Selective shrinkage}\label{sec:selective_shrinkage}
By ``selective shrinkage'' we mean that the degree of shrinkage
applied to a collection of estimators varies across estimators. As
motivated in Section~\ref{sec:gpc_and_shrinkage}, we are specifically
interested in selectively shrinking posterior distributions near
isolated locations more strongly than in dense regions. This section shows
that by changing the \emph{form} of prior marginals (heavy-tailed
instead of Gaussian) we can induce stronger selective shrinkage than
any GPR. Since HPR uses a GP in its construction, which can induce
(some) selective shrinkage on its own, care must be taken to
investigate only the additional benefits the
transformation~\(\GinvPhi{\cdot}\) has on shrinkage. For this reason
we assume a particular GP prior which leads to a special type of
shrinkage in GPR and then check how an HPR model built on top of that
GP changes the observed behavior.

In this section we provide an idealized analysis of that allows us to
compare the selective shrinkage obtained by GPR and HPR.  Note that we
focus on regression in this section so that we can obtain analytical
results.  We work with \(n\) measurement locations, \(X = \left(x_1,
\ldots, x_n\right)\), whose index set \(\{1, \ldots, n\}\) can be
partitioned into a ``dense'' set \(D\) with \(|D| = n-1\) and a single
``sparse'' index \(s \notin D\). Assume that \(x_d = x_{d'}\ \forall
d,d' \in D\) so that we may let (without loss of generality) \(\tilde
K(x_d,x_{d'}) = 1\ \forall d\neq d' \in D\). We also assert that \(x_d
\neq x_s\ \forall d \in D\) and let \(\tilde K(x_d, x_s) = \tilde
K(x_s, x_d) = 0\ \forall d \in D\). Assuming that \(n > 2\) we fix the
remaining entry \(\tilde K(x_s,x_s) = \epsilon / (\epsilon + n - 2)\),
for some \(\epsilon > 0\). We interpret \(\epsilon\) as a noise
variance and let \(K = \tilde K + \epsilon I\). The set of locations
\(X\) idealizes an uneven sampling of input space, consisting of a
densely and a sparsely sampled region, as represented by \(D\) and
\(s\).

Denote any distributions computed under the GPR model by
\(p_{\mathrm{gp}}(\cdot)\) and those computed in HPR by
\(p_{\mathrm{hp}}(\cdot)\). Using \(K(X, X) = K\), define \(z(X)\) as in
Eq.~\eqref{eq:const0}.  Let \(y\) denote a vector of real-valued
measurements for a regression task.  The posterior distribution of
\(z(x_i)\) given \(y\), with \(x_i \in X\), is derived by standard
Gaussian computations as
\begin{align*}
  &p_{\mathrm{gp}}(z(x_i)|X, y) = \mathcal{N}\left(\mu_i, \sigma_i^2\right) \\
  &\mu_i = \tilde K(x_i, X) K(X, X)^{-1} y\\
  &\sigma_i^2 = K(x_i,x_i) - \tilde K(x_i,X) K(X, X)^{-1}
  \tilde K(X,x_i).
\end{align*}
For our choice of \(K(X, X)\) one can show that \(\sigma_{d}^2 =
\sigma_{s}^2\) for \(d \in D\).  To ensure that the posterior
distributions agree at the two locations we require \(\mu_{d} =
\mu_{s}\), which holds if measurements \(y\) satisfy
\begin{align*}
  y \in \mathcal{Y}_{\mathrm{gp}} &\triangleq \left\{y | \left(\tilde K(x_d,
  X) - \tilde K(x_s, X)\right) K(X, X)^{-1} y = 0\right\} = \left\{y \left| \sum_{d\in D}y_d = y_s\right.\right\}.
\end{align*}
\begin{figure}
  \centering
  \subfigure[\tiny \(g_b(x) = \frac{1}{2b}\exp\left\{-\frac{|x|}{b}\right\}\)]{ 
    \includegraphics[width=4.2cm]{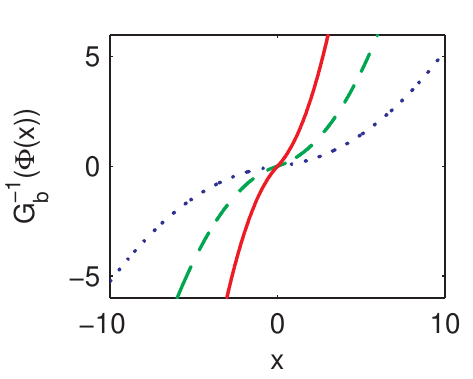} 
    \label{fig:GinvPhi_lap}
  }
  \subfigure[\tiny \(g_b(x) = \frac{1}{2b} \mathrm{sech}\left(\frac{\pi x}{2b}\right)\)]{
    \includegraphics[width=4.2cm]{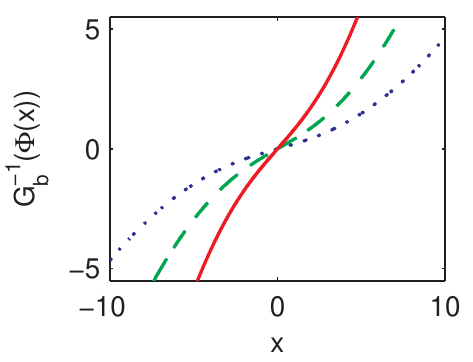} 
    \label{fig:GinvPhi_hypsec}
  }
  \subfigure[\tiny \(g_b(x) = \frac{1}{b\left(2 + (x/b)^2\right)^{3/2}}\)]{
    \includegraphics[width=4.2cm]{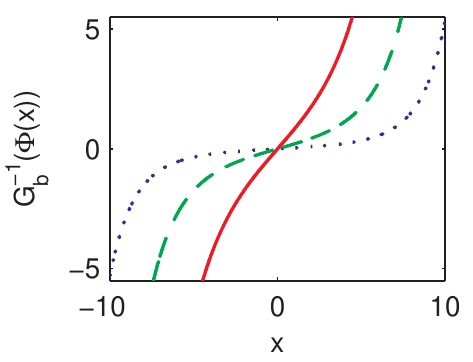} 
    \label{fig:GinvPhi_studentt}
  }
  \caption{Illustration of \(\GinvPhi{x}\), for \(\sigma^2 = 1.0\)
    with \(G_b\) the c.d.f.\ of~\subref{fig:GinvPhi_lap} the Laplace
    distribution~\subref{fig:GinvPhi_hypsec} the hyperbolic
    secant distribution~\subref{fig:GinvPhi_studentt} a Student-\(t\)
    inspired distribution, all with scale parameter \(b\). Each plot
    shows three samples---dotted, dashed, solid---for growing \(b\).
    As \(b\) increases the distributions become heavy-tailed and the
    gradient of \(\GinvPhi{x}\) increases.}\label{fig:GinvPhi}
\end{figure}
A similar analysis can be carried out for the induced HPR model. By
Eqs.~\eqref{eq:zspace_pred_start}--\eqref{eq:zspace_pred_end} HPR
inference leads to identical distributions \(p_{\mathrm{hp}}(z(x_d)|X,
y') = p_{\mathrm{hp}}(z(x_s)|X, y')\) with \(d \in D\) if measurements
\(y'\) in \(f\)-space satisfy
\begin{align*}
  y' \in \mathcal{Y}_{\mathrm{hp}} &\triangleq \left\{y' |
  \left(\tilde K(x_d, X) - \tilde K(x_s, X)\right) K(X, X)^{-1}\PhiinvG{y'} = 0\right\} \\ &= \left\{y' = \GinvPhi{y} | y \in \mathcal{Y}_{\mathrm{gp}}\right\}.
\end{align*}
To compare the shrinkage properties of GPR and HPR we analyze select
pairs of measurements in \(\mathcal{Y}_{\mathrm{gp}}\) and
\(\mathcal{Y}_{\mathrm{hp}}\). The derivation requires that
\(\GinvPhi{\cdot}\) is strongly concave on \((-\infty, 0]\), strongly
  convex on \([0, +\infty)\) and has gradient \(> 1\) on
    \(\mathbb{R}\). To see intuitively why this should hold for
    heavy-tailed marginals, note that for \(G_b\) with fatter tails
    than a Gaussian, \(|\GinvPhi{x}|\) should eventually dominate
    \(|\Phi_{0, b^2}^{-1}(\Phi_{0, \sigma^2}(x))| = (b/\sigma) |x|\).
    Figure~\ref{fig:GinvPhi} demonstrates graphically that the
    assumption holds for several choices of \(G_b\), provided \(b\) is
    large enough, i.e., that \(g_b\) has sufficiently heavy
    tails. Indeed, it can be shown that for scale parameters \(b >
    0\), the first and second derivatives of \(\GinvPhi{\cdot}\) scale
    linearly with \(b\). Consider a measurement \(0 \neq y \in
    \mathcal{Y}_{\mathrm{gp}}\) with \(\sign{y(x_d)} =
    \sign{y(x_{d'})}, \forall d, d' \in D\). Analyzing such \(y\) is
    relevant, as we are most interested in comparing how multiple
    \emph{reinforcing} observations at clustered locations and a
    single isolated observation are absorbed during inference. By
    definition of \(\mathcal{Y}_{\mathrm{gp}}\), for \(d^* =
    \argmax_{d\in D} |y_d|\) we have \(|y_{d^*}| < |y_s|\) as long as
    \(n > 2\). The corresponding element \(y' = \GinvPhi{y} \in
    \mathcal{Y}_{\mathrm{hp}}\) then satisfies
    \begin{align}
      |y'(x_s)| = \left|\GinvPhi{y(x_s)}\right| > \left|\frac{\GinvPhi{y(x_{d^*})}}{y(x_{d^*})} y(x_s)\right| = \left|\frac{y'(x_{d^*})}{y(x_{d^*})} y(x_s)\right|.\label{eq:shrinkage_inequality}
    \end{align}
Thus HPR inference leads to identical predictive distributions in
\(f\)-space at the two locations even though the isolated observation
\(y'(x_s)\) has disproportionately larger magnitude than
\(y'(x_{d^*})\), relative to the GPR measurements \(y(x_s)\) and
\(y(x_{d^*})\). As this statement holds for any \(y\in
\mathcal{Y}_{\mathrm{gp}}\) satisfying our earlier sign requirement,
it indicates that HPR systematically shrinks isolated observations
more strongly than GPR. Moreover, since the second derivative of
\(\GinvPhi{\cdot}\) scales linearly with \(b > 0\) an intuitive
connection suggests itself when looking at
inequality~\eqref{eq:shrinkage_inequality}: the heavier the marginal
tails, the stronger the inequality and thus the stronger the selective
shrinkage.

The previous derivation exemplifies in an idealized setting that HPR
leads to improved shrinkage of predictive distributions near isolated
observations. More generally, because GPR transforms measurements only
linearly, while HPR additionally pre-transforms measurements
nonlinearly, our analysis suggests that for any GPR we can find an HPR
model which leads to stronger selective shrinkage. The result has
intuitive parallels to the parametric case: just as
\(\ell_1\)-regularization improves shrinkage of parametric estimators,
heavy-tailed processes improve shrinkage of nonparametric
estimators. Although our analysis kept \(K(X, X)\) fixed for
GPR and HPR, in practice we are free to tune the kernel to yield a
desired scale of predictive distributions.  Lastly, although our
analysis has been carried out for regression, it motivates us to
explore heavy-tailed processes in the classification setting.
\section{Heavy-tailed process classification}\label{sec:sparse_proc_classification}
The derivation of our \emph{heavy-tailed process classifier} (HPC) is similar
to that of multiclass GPC with Laplace approximation in Section 3.5 of
Rasmussen and Williams~\cite{rasmussenwilliams}. However, due to the
nonlinear transformations involved, some nice properties of their
derivation are lost. We revert notation and let \(y\) denote a vector
of class labels. For a \(C\)-class classification problem with \(n\)
training points we introduce a vector of \(nC\) latent function
measurements
\begin{align*}
  f = (f_1^1, \ldots, f_n^1, f_1^2, \ldots, f_n^2, \ldots,
  f_1^C, \ldots, f_n^C)^\top.
\end{align*}
For each block \(c \in \{1, \ldots, C\}\) of \(n\) variables we define
an independent heavy-tailed process prior using
Eq.~\eqref{eq:warped_likelihood} with a kernel matrix \(K_c\).
Equivalently, we can define the prior jointly on \(f\) by letting
\(K\) be a block-diagonal kernel matrix with blocks \(K_1, \ldots,
K_C\). Each kernel matrix \(K_c\) is defined by a (possibly different)
symmetric positive semidefinite kernel with its own set of
parameters. The following construction relaxes the earlier condition
that \(\diag{K} = \sigma^2{\bf 1}\) and instead views \(\Phi_{0,
  \sigma^2}(\cdot)\) as just some nonlinear transformation with
parameter \(\sigma^2\). By this relaxation we effectively adopt Liu et
al.'s~\cite{liu2009nonparanormal} interpretation that
Eq.~\eqref{eq:warped_likelihood} defines the copula. The scale
parameters \(b\) could in principle vary across the \(nC\) variables,
but we keep them constant at least within each block of \(n\).  Labels
\(y\) are represented in a 1-of-$n$ form and generated by the following
observation model
\begin{align}
  p(y_i^c = 1|f_i) = \pi_i^c =
  \frac{\exp\{f_i^c\}}{\sum_{c'}\exp\{f_i^{c'}\}}.\label{eq:spc_likelihood}
\end{align}
For inference we are ultimately interested in computing 
\begin{align}
  p(y_*^c = 1|X, y, x_*) &= \mathbb{E}_{p(f_*| X, y, x_*)}\left(\frac{\exp\{f_*^c\}}{\sum_{c'}\exp\{f_*^{c'}\}}\right),\label{eq:spc_predictive_prob}
\end{align}
where \(f_* = (f_*^1, \ldots, f_*^C)^\top\). The previous section
motivates the hope that improved selective shrinkage will occur in \(p(f_*|X,
y, x_*)\), provided the prior marginals have sufficiently heavy tails.
\subsection{Inference}
As in GPC, most of the intractability lies in computing the predictive
distribution \(p(f_*|X, y, x_*)\). We use the Laplace approximation to
address this issue: a Gaussian approximation to \(p(z|X, y)\) is found
and then combined with the Gaussian \(p(z_*|X, z, x_*)\) to give us an
approximation to \(p(z_*|X, y, x_*)\). This is then transformed to a
(typically non-Gaussian) distribution in \(f\)-space using a change of
variables.  Hence we first seek to find a mode and corresponding
Hessian matrix of the log posterior \(\log p(z|X, y)\). Recalling the
relation \(f = \GinvPhi{z}\), the log posterior can be written as
\begin{align*}
  J(z) &\triangleq \log p(y|z) + \log p(z)\\
  &= y^\top f -
  \sum_i\log\sum_c\exp\left\{f_i^c)\right\}
  -\frac{1}{2}z^\top K^{-1}z - \frac{1}{2}\log |K| + \mathrm{const.}
\end{align*}
Let \(\Pi\) be an \(nC\times n\) matrix of stacked diagonal matrices
\(\diag{\pi^c}\) for \(n\)-subvectors \(\pi^c\) of \(\pi\). With \(W =
\diag{\pi} - \Pi\Pi^\top\), the gradients are
\begin{align*}
  \nabla J(z) &= \diag{\frac{df}{dz}}(y - \pi) - K^{-1}z\\
  \nabla^2 J(z) &= \diag{\frac{d^2f}{dz^2}}\diag{y -
    \pi} - \diag{\frac{df}{dz}}W\diag{\frac{df}{dz}} - K^{-1}.
\end{align*}
Unlike in Rasmussen and Williams~\cite{rasmussenwilliams}, \(-\nabla^2
J(z)\) is not generally positive definite owing to its first term. For
that reason we cannot use a Newton step to find the mode and instead
resort to a simpler gradient method. Once the mode \(\hat z\) has been
found we approximate the posterior as
\begin{align*}
  p(z|X, y) \approx q(z|X, y) = \mathcal{N}\left(\hat z, -\nabla^2 J(\hat z)^{-1}\right),
\end{align*}
and use this to approximate the predictive distribution by
\begin{align*}
  q(z_*|X, y, x_*) = \int p(z_*|X, z, x_*) q(z|X, y)df.
\end{align*}
Since we arranged for both distributions in the integral to be
Gaussian, the resulting Gaussian can be straightforwardly
evaluated. Finally, to approximate the one-dimensional integral with
respect to \(p(f_*|X, y, x_*)\) in
Eq.~\eqref{eq:spc_predictive_prob} we could either use a
quadrature method, or generate samples from \(q(z_*|X, y, x_*)\),
convert them to \(f\)-space using \(\GinvPhi{\cdot}\) and then
approximate the expectation by an average. We have compared
predictions resulting using the latter method with those of a Gibbs
sampler; the Laplace approximation matched Gibbs results well, while
costing only a fraction of time to compute.
\subsection{Parameter estimation}
Using a derivation similar to that in section 3.4.4
of~\cite{rasmussenwilliams}, we have for \(\hat f = \GinvPhi{\hat z}\)
that the Laplace approximation of the marginal log likelihood is
\begin{align}
  \log p(y|x) &\approx \log q(y|x) = J(\hat z) - \frac{1}{2} \log
  |-2\pi\nabla^2 J(\hat z)|\label{eq:lap_log_marg_approx}\\
  &= y^\top\hat f - \sum_i\log\sum_c\exp\left\{\hat
    f_i^c\right\} - \frac{1}{2}\hat z^\top K^{-1} \hat z -
  \frac{1}{2}\log|K| - \frac{1}{2}\log|-\nabla^2 J(\hat z)| + \mathrm{const.}\nonumber
\end{align}
We optimize kernel parameters \(\theta\) by taking gradient steps on
\(\log q(y|x)\). The derivative needs to take into account that perturbing the
parameters can also perturb the mode \(\hat z\) found for the Laplace
approximation. At an optimum \(\nabla J(\hat z)\) must be zero, so
that
\begin{align}
  \hat z = K\diag{\frac{d\hat f}{d\hat z}}(y - \hat \pi),\label{eq:mode}
\end{align}
where \(\hat \pi\) is defined as in Eq.~\eqref{eq:spc_likelihood}
but using \(\hat f\) rather than \(f\).  Taking derivatives of this
equation allows us to compute the gradient \(d\hat z/
d\theta\). Differentiating the marginal likelihood we have
\begin{align*}
  \frac{d\log q(y|x)}{d\theta} &= (y-\hat \pi)^\top \diag{\frac{d\hat f}{d\hat z}}
  \frac{d\hat z}{d\theta} - \frac{d\hat z}{d\theta}K^{-1}\hat z + \frac{1}{2}\hat z^\top
  K^{-1} \frac{dK}{d\theta}K^{-1}\hat z~-\\ 
  &~~~~~\frac{1}{2}\mathrm{tr}\left(K^{-1}\frac{dK}{d\theta}\right) -
  \frac{1}{2}\mathrm{tr}\left(\nabla^2 J(\hat
  z)^{-1}\frac{d\nabla^2 J(\hat z)}{d\theta}\right).
\end{align*} 
\begin{figure}
  \centering
  \subfigure[]{
    \raisebox{1cm}{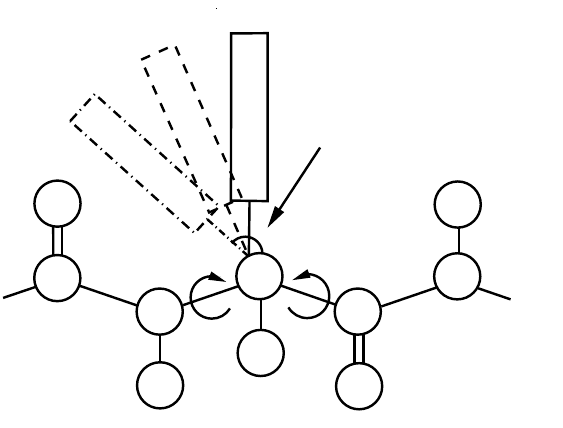}
    \label{fig:rotamer}
  }
  \subfigure[]{ 
    \includegraphics[width=6cm]{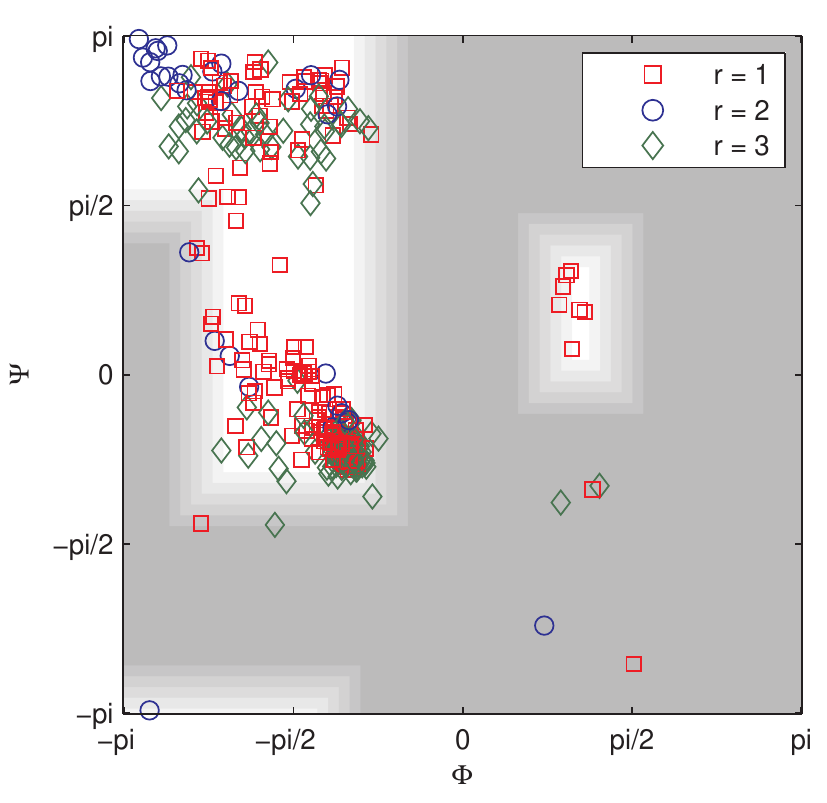}
    \label{fig:ramachandran}
  }\caption{\subref{fig:rotamer} Schematic of a protein section. The
    backbone is the sequence of \(C', N, C_\alpha, C', N\) atoms. An
    amino-acid-specific sidechain extends from the \(C_\alpha\) atom
    at one of three discrete angles known as
    ``rotamers.'' \subref{fig:ramachandran} Ramachandran plot of \(400\)
    \((\Phi, \Psi)\) measurements and corresponding rotamers (by
    shapes/colors) for amino-acid {\tt arg}. The dark shading
    indicates the sparse region we considered in producing results in
    Figure~\ref{fig:results}. Progressively lighter shadings indicate
    how the sparse region was grown to produce
    Figure~\ref{fig:evolution}.}
\end{figure} 
The remaining gradient computations are straightforward, albeit
tedious. In addition to optimizing the kernel parameters, it may also
be of interest to optimize the scale parameter \(b\) of marginals
\(G_b\). Again, differentiating Eq.~\eqref{eq:mode} with respect
to \(b\) allows us to compute \(d\hat z/db\).  We note that when
perturbing \(b\) we change \(\hat f\) by changing the underlying mode
\(\hat z\) as well as by changing the parameter \(b\) which is used to
compute \(\hat f\) from \(\hat z\). Suppressing the detailed
computations, the derivative of the marginal log likelihood with
respect to \(b\) is
\begin{align*}
  \frac{d\log q(y|x)}{db} &= (y-\hat \pi)^\top\frac{d\hat f}{db}
  - \frac{d\hat z}{db}^\top K^{-1} \hat z -
  \frac{1}{2}\mathrm{tr}\left(\nabla^2 J(\hat z)^{-1}
  \frac{d\nabla^2 J(\hat z)}{db}\right).
\end{align*}
\section{Experiments}\label{sec:experiments}
To a first approximation, the three-dimensional structure of a folded
protein is defined by pairs of continuous backbone angles \((\Phi,
\Psi)\), one pair for each amino-acid, as well as discrete angles,
so-called rotamers, that define the conformations of the amino-acid
sidechains that extend from the backbone. The geometry is outlined in
Figure~\ref{fig:rotamer}. There is a strong dependence between
backbone angles \((\Phi, \Psi)\) and rotamer values; this is
illustrated in the ``Ramachandran plot'' shown in
Figure~\ref{fig:ramachandran}, which plots the backbone angles for
each rotamer (indicated by the shapes/colors). The dependence is
exploited in computational approaches to protein structure prediction,
where estimates of rotamer probabilities given backbone angles are
used as one term in an energy function that models native protein
states as minima of the energy. Protein structures are predicted by
minimizing the energy function. Poor estimates of rotamer
probabilities in sparse regions can derail the prediction
procedure. Indeed, sparsity has been a serious problem in
state-of-the-art rotamer models based on kernel density estimates
(Roland Dunbrack, personal communication). Unfortunately, we have
found that GPC is not immune to the sparsity problem.

\begin{figure}
  \centering
  \includegraphics[width=6.5cm]{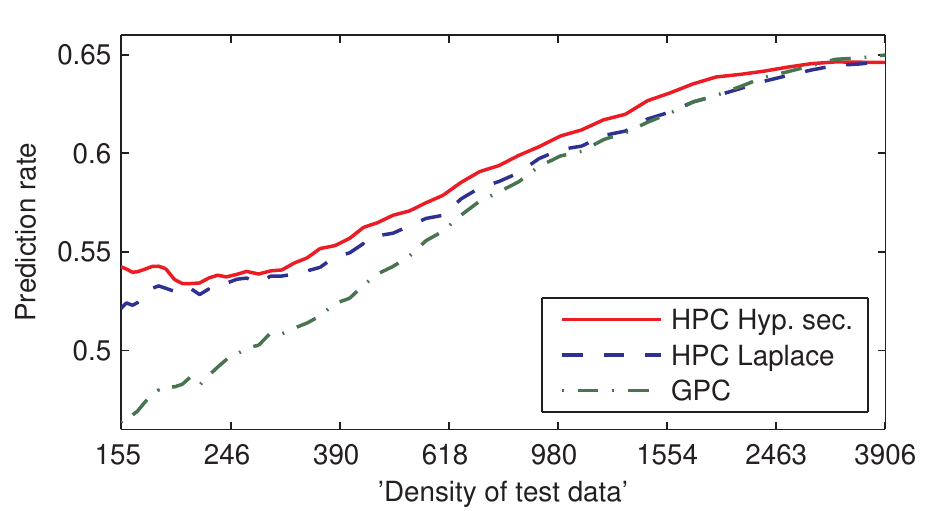}
  \caption{Average rotamer prediction rate in the sparse region for
    both flavors of HPC as well as standard GPC as a function of the average
    number of points in the sparse region.}
  \label{fig:evolution}
\end{figure}

To evaluate our algorithm we consider rotamer-prediction tasks on the
17 amino-acids (out of 20) that have three rotamers at the first
dihedral angle along the sidechain\footnote{Residues {\tt ala} and
  {\tt gly} are non-discrete while {\tt pro} has only two rotamers at
  the first dihedral angle.}. Our previous work thus applies with the
number of classes \(C = 3\) and the covariates being \((\Phi,
\Psi)\) angle pairs.  Since the input space is a torus we defined GPC
and HPC using the following von Mises-inspired kernel for
\(d\)-dimensional angular data:
\begin{align*}
  k(x_i, x_j) = \sigma^2 \exp\left\{\lambda \left(\left(\sum_{k=1}^d
  \cos(x_{i,k} - x_{j,k})\right) - d\right)\right\},
\end{align*}
where \(x_{i,k}, x_{j, k} \in [0, 2\pi]\) and \(\sigma^2, \lambda \ge
0\)\footnote{The function \(\cos(x_{i,k} - x_{j, k}) = [\cos(x_{i.k}),
    \sin(x_{i,k})] [\cos(x_{j.k}), \sin(x_{j,k})]^\top\) is a
  symmetric positive semi-definite kernel. By Propositions 3.22 (i)
  and (ii) and Proposition 3.25 in Shawe-Taylor and
  Cristianini~\cite{shawetaylor2004kernelmethods}, so is \(k(x_i,
  x_j)\) above.}.  To find good GPC kernel parameters we optimize an
\(\ell_2\)-regularized version of the Laplace approximation to the log
marginal likelihood reported in Eq.~3.44
of~\cite{rasmussenwilliams}. For HPC we let \(G_b\) be either the
centered Laplace distribution or the hyperbolic secant distribution
with scale parameter \(b\). We estimate HPC kernel parameters as well
as \(b\) by similarly maximizing an \(\ell_2\)-regularized form of
Eq.~\eqref{eq:lap_log_marg_approx}.  In both cases we restricted the
algorithms to training sets of only 100 datapoints. Since good
regularization parameters for the objectives are not known a priori we
train with and test them on a grid for each of the 17 rotameric
residues in ten-fold cross-validation. To find good regularization
parameters for a particular residue we look up that combination which,
averaged over the ten folds of the remaining 16 residues, produced the
best test results. Having chosen the regularization constants we
report average test results computed in ten-fold cross validation.
\begin{figure}
  \centering
  \subfigure[]{ 
    \includegraphics[width=6.7cm]{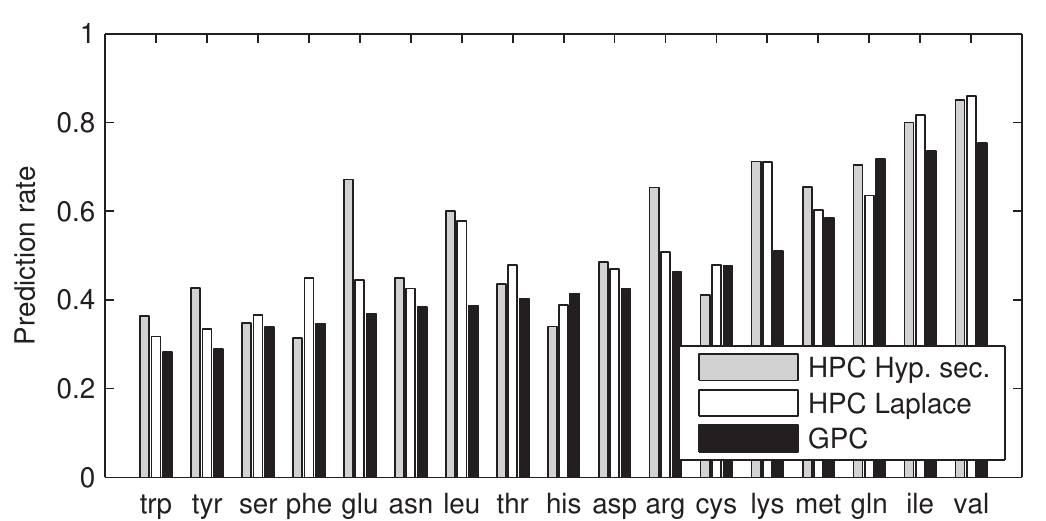}
    \label{fig:sparse_results}
  }
  \subfigure[]{ 
    \includegraphics[width=6.7cm]{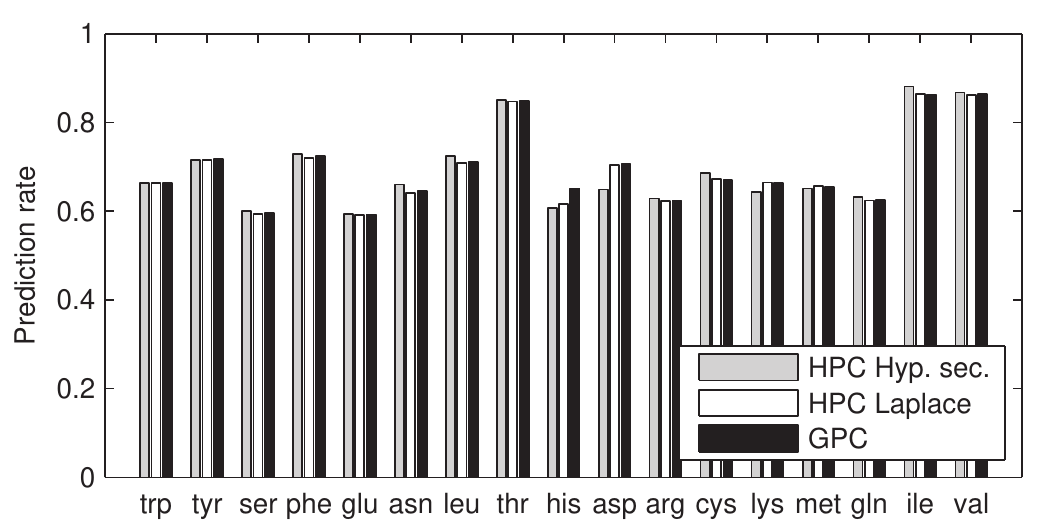}
    \label{fig:dense_results}
  }
  \caption{Rotamer prediction rates in percent
    in~\subref{fig:sparse_results} sparse
    and~\subref{fig:dense_results} dense regions. Both flavors of HPC
    (hyperbolic secant and Laplace marginals) significantly outperform GPC in
    sparse regions while performing competitively in dense regions.}
  \label{fig:results}
\end{figure}
We evaluate the algorithms on predefined sparse and dense regions in
the Ramachandran plot, as indicated by the background shading in
Figure~\ref{fig:ramachandran}. Across 17 residues the sparse regions
usually contained more than 70 measurements (and often more than 150),
each of which appears in one of the 10 cross-validation
folds. Figure~\ref{fig:results} compares the label prediction rates on
the dense and sparse regions. Averaged over all 17 residues HPC
outperforms GPC by 5.79\% with Laplace and 7.89\% with hyperbolic
secant marginals. With Laplace marginals HPC underperforms GPC on only
two residues in sparse regions: by 8.22\% on {\tt gln}, and by 2.53\%
on {\tt his}. On dense regions HPC lies within 0.5\% on 16 residues
and only degrades once by 3.64\% on {\tt his}. Using hyperbolic secant
marginals HPC often improves GPC by more than 10\% on sparse regions
and degrades by more than 5\% only on {\tt cys} and {\tt his}. On
dense regions HPC usually performs within 1.5\% of GPC. In
Figure~\ref{fig:evolution} we show how the average rotamer prediction
rate across 17 residues changes as we grow the sparse region to
include more measurements from dense regions. The growth of the sparse
region is indicated by progressively lighter shadings in
Figure~\ref{fig:ramachandran}. As more points are included the
significant advantage of HPC lessens. Eventually GPC does marginally
better than HPC. The values reported in Figure~\ref{fig:results}
correspond to the dark shaded region, which contains an average of 155
measurements per residue.
\section{Related research}\label{sec:related}
Copulas~\cite{copulabook} allow convenient modelling of multivariate
correlation structures as separate from marginal distributions. Early
work by Song~\cite{multivariatedisp2000song} used the Gaussian copula
to generate complex multivariate distributions by complementing a
simple copula form with  marginal distributions of
choice. Popularity of the Gaussian copula in the financial literature
is generally credited to Li~\cite{li2000copula} who used it to model
correlation structure for pairs of random variables with known
marginals. More recently, the Gaussian process has been modified in a
similar way to ours by Snelson et al.~\cite{snelson04warped} who
called the resulting stochastic process a Warped Gaussian
Process. They demonstrate that posterior distributions can better
approximate the true noise distribution if the transformation defining
the warped process is learned.  Jaimungal and Ng~\cite{jaimungal09kcp}
have extended this work to model multiple parallel time series with
marginally non-Gaussian stochastic processes under the name of
Kernel-based Copula Processes (KCPs). Their work uses a ``binding
copula'' to combine several subordinate copulas into a joint
model. Bayesian approaches focusing on estimation of the Gaussian
copula covariance matrix for a given dataset are given
in~\cite{dobra2009copula, pitt2006gaussiancopularegression}. With the
advent of larger datasets, research has also focused on estimation in
high-dimensional settings. Liu et al.~\cite{liu2009nonparanormal} do
away with a prior on covariance matrices and give consistency
results for a covariance estimator in high-dimensional
settings.
\section{Conclusions}\label{sec:conclusion}
This paper has analyzed learning scenarios where outliers are observed
in the input space, rather than the output space as commonly discussed
in the literature. We illustrated heavy-tailed processes as a
straightforward extension of GPs and an elegant and economical way to
improve the robustness of estimators in sparse regions beyond those of
GP-based methods. This was demonstrated both by a theoretical analysis
and experimental results. Importantly, because heavy-tailed processes
are based on a GP, they inherit many of its favorable computational
properties; predictive inference in regression, for instance, is
straightforward. For approximate inference in more complicated models
utilizing heavy-tailed processes, this paper exemplifies that we can
borrow many ideas from standard GP models. Since heavy-tailed
processes have a parsimonious representation, they can be easily used
as building blocks in more complicated models where currently GPs are
used. The benefits of heavy-tailed processes on selective shrinkage
thus extend to many other GP-based models that currently struggle with
covariate shift.
\section{Acknowledgements}
We would like to thank Roland Dunbrack and Maxim Shapovalov for
helpful discussions and for giving us access to the data.
\bibliography{bibliography}
\end{document}

%% file: rotamer_schematic_sidechain.pdf_t
\begin{picture}(0,0)%
\includegraphics{rotamer_schematic_sidechain.pdf}%
\end{picture}%
\setlength{\unitlength}{1533sp}%
\begingroup\makeatletter\ifx\SetFigFont\undefined%
\gdef\SetFigFont#1#2#3#4#5{%
  \reset@font\fontsize{#1}{#2pt}%
  \fontfamily{#3}\fontseries{#4}\fontshape{#5}%
  \selectfont}%
\fi\endgroup%
\begin{picture}(6917,5237)(3090,-5592)
\put(3682,-2944){\makebox(0,0)[lb]{\smash{{\SetFigFont{7}{8.4}{\familydefault}{\mddefault}{\updefault}{\color[rgb]{0,0,0}$O$}%
}}}}
\put(4925,-5187){\makebox(0,0)[lb]{\smash{{\SetFigFont{7}{8.4}{\familydefault}{\mddefault}{\updefault}{\color[rgb]{0,0,0}$H$}%
}}}}
\put(4906,-4283){\makebox(0,0)[lb]{\smash{{\SetFigFont{7}{8.4}{\familydefault}{\mddefault}{\updefault}{\color[rgb]{0,0,0}$N$}%
}}}}
\put(8582,-3847){\makebox(0,0)[lb]{\smash{{\SetFigFont{7}{8.4}{\familydefault}{\mddefault}{\updefault}{\color[rgb]{0,0,0}$N$}%
}}}}
\put(8602,-2953){\makebox(0,0)[lb]{\smash{{\SetFigFont{7}{8.4}{\familydefault}{\mddefault}{\updefault}{\color[rgb]{0,0,0}$H$}%
}}}}
\put(7357,-4285){\makebox(0,0)[lb]{\smash{{\SetFigFont{7}{8.4}{\familydefault}{\mddefault}{\updefault}{\color[rgb]{0,0,0}$C'$}%
}}}}
\put(7408,-5205){\makebox(0,0)[lb]{\smash{{\SetFigFont{7}{8.4}{\familydefault}{\mddefault}{\updefault}{\color[rgb]{0,0,0}$O$}%
}}}}
\put(6753,-1965){\makebox(0,0)[lb]{\smash{{\SetFigFont{9}{10.8}{\familydefault}{\mddefault}{\updefault}{\color[rgb]{0,0,0}Rotamer $r \in \{1, 2, 3\}$}%
}}}}
\put(7078,-3758){\makebox(0,0)[lb]{\smash{{\SetFigFont{11}{13.2}{\familydefault}{\mddefault}{\updefault}{\color[rgb]{0,0,0}$\Phi$}%
}}}}
\put(6168,-4783){\makebox(0,0)[lb]{\smash{{\SetFigFont{7}{8.4}{\familydefault}{\mddefault}{\updefault}{\color[rgb]{0,0,0}$H$}%
}}}}
\put(3630,-3878){\makebox(0,0)[lb]{\smash{{\SetFigFont{7}{8.4}{\familydefault}{\mddefault}{\updefault}{\color[rgb]{0,0,0}$C'$}%
}}}}
\put(5221,-3751){\makebox(0,0)[lb]{\smash{{\SetFigFont{11}{13.2}{\familydefault}{\mddefault}{\updefault}{\color[rgb]{0,0,0}$\Psi$}%
}}}}
\put(6088,-3802){\makebox(0,0)[lb]{\smash{{\SetFigFont{7}{8.4}{\familydefault}{\mddefault}{\updefault}{\color[rgb]{0,0,0}$C_\alpha$}%
}}}}
\put(4409,-2030){\rotatebox{318.0}{\makebox(0,0)[lb]{\smash{{\SetFigFont{7}{8.4}{\familydefault}{\mddefault}{\updefault}{\color[rgb]{0,0,0}Sidechain}%
}}}}}
\put(5159,-1435){\rotatebox{294.0}{\makebox(0,0)[lb]{\smash{{\SetFigFont{7}{8.4}{\familydefault}{\mddefault}{\updefault}{\color[rgb]{0,0,0}Sidechain}%
}}}}}
\put(6079,-1210){\rotatebox{270.0}{\makebox(0,0)[lb]{\smash{{\SetFigFont{7}{8.4}{\familydefault}{\mddefault}{\updefault}{\color[rgb]{0,0,0}Sidechain}%
}}}}}
\end{picture}%